\documentclass[runningheads]{llncs}

\usepackage{amsmath}
\usepackage{amsfonts}
\usepackage{amssymb}

\usepackage{paralist}
\usepackage{booktabs} 

\usepackage{esvect}
\usepackage{multirow}
\usepackage{subcaption}
\usepackage[para,online,flushleft]{threeparttable}

\usepackage{epstopdf}
\usepackage{graphics}
\usepackage{graphicx}
\DeclareGraphicsExtensions{.eps,.gz,.eps,.pdf,.png,.jpg}
\graphicspath{{./}{../}}

\usepackage[utf8]{inputenc}
\usepackage[tablename=Table]{caption} 
\usepackage[figurename=Figure]{caption}
\usepackage{upgreek}
\usepackage{array}
\usepackage{graphicx}
\usepackage{lineno,hyperref}
\modulolinenumbers[5]
\usepackage{float}
\usepackage{multirow}

\usepackage{algorithm}
\usepackage{algorithmicx}
\usepackage{algpseudocode}
\usepackage{adjustbox}

\usepackage{bbm}
\usepackage{dsfont}

\usepackage{xcolor}
\usepackage{color, soul}
\usepackage{listings}

\newcommand{\sstitle}[1]{\smallskip\noindent\textbf{#1.\/}}

\def\Snospace~{\S{}}

\makeatletter
\newcommand{\removelatexerror}{\let\@latex@error\@gobble}
\makeatother

\colorlet{punct}{red!60!black}
\definecolor{background}{HTML}{EEEEEE}
\definecolor{delim}{RGB}{20,105,176}
\colorlet{numb}{magenta!60!black} 

\lstdefinelanguage{json}{
    basicstyle=\normalfont\ttfamily,
    numbers=left,
    numberstyle=\scriptsize,
    stepnumber=1,
    numbersep=8pt,
    showstringspaces=false,
    breaklines=true,
    frame=lines,
    backgroundcolor=\color{background},
    literate=
     *{0}{{{\color{numb}0}}}{1}
      {1}{{{\color{numb}1}}}{1}
      {2}{{{\color{numb}2}}}{1}
      {3}{{{\color{numb}3}}}{1}
      {4}{{{\color{numb}4}}}{1}
      {5}{{{\color{numb}5}}}{1}
      {6}{{{\color{numb}6}}}{1}
      {7}{{{\color{numb}7}}}{1}
      {8}{{{\color{numb}8}}}{1}
      {9}{{{\color{numb}9}}}{1}
      {:}{{{\color{punct}{:}}}}{1}
      {,}{{{\color{punct}{,}}}}{1}
      {\{}{{{\color{delim}{\{}}}}{1}
      {\}}{{{\color{delim}{\}}}}}{1}
      {[}{{{\color{delim}{[}}}}{1}
      {]}{{{\color{delim}{]}}}}{1},
}

\begin{document}




\title {A Comparative Study of Question Answering over Knowledge Bases}  

\author{
Khiem Vinh Tran\inst{1}
\and
Hao Phu Phan\inst{2}
\and
Khang Nguyen Duc Quach\inst{3}
\and
Ngan Luu-Thuy Nguyen\inst{1}
\and
Jun Jo\inst{3}
\and
Thanh Tam Nguyen\inst{3}
}

\authorrunning{Khiem Vinh Tran et al.} 

\institute{
University of Information Technology, Ho Chi Minh City, Vietnam
\and
HUTECH University, Ho Chi Minh City, Vietnam
\and
Griffith University, Gold Coast, Australia
}

%
%
%
%

\maketitle

\abstract{
Question answering over knowledge bases (KBQA) has become a popular approach to help users extract information from knowledge bases. Although several systems exist, choosing one suitable for a particular application scenario is difficult. In this article, we provide a comparative study of six representative KBQA systems on eight benchmark datasets. In that, we study various question types, properties, languages, and domains to provide insights on where existing systems struggle. On top of that, we propose an advanced mapping algorithm to aid existing models in achieving superior results. Moreover, we also develop a multilingual corpus COVID-KGQA, which encourages COVID-19 research and multilingualism for the diversity of future AI. Finally, we discuss the key findings and their implications as well as performance guidelines and some future improvements. Our source code is available at \url{https://github.com/tamlhp/kbqa}.

\keywords{Question answering  \and Knowledge base \and Query processing}
}






\section{Introduction}

Question Answering (QA) is a long-standing discipline within the field of natural language processing (NLP), which is concerned with providing answers to questions posed in natural language on data sources, and also draws on techniques from linguistics, database processing, and information retrieval~\cite{ADMA3,hung2013evaluation,nguyen2015result,tam2019anomaly,nguyen2020monitoring}. 
One important type of data sources is knowledge bases, also known as knowledge graphs, which have been automatically constructed
from web data and have become a key asset for search engines
and many applications
\cite{Weikum2021}. 
Finding answers for a question in a KB, on the other hand, is not always straightforward. The user needs to have a thorough understanding of the KB as well as a structured query language in order to express their queries in a structured manner that can be utilized to locate matches in the KB. 

In order to address this issue, a significant number of QA systems that allow users to express their information requirements using natural language have been created. Additionally, factoid question answering has two main approaches, information retrieval (IR) based QA and knowledge-based QA. In the first approach, many research have been established in recent years with the advancement of machine reading comprehension (MRC) task. The second approach is question answering over knowledge base (KBQA) with precision of question answering over knowledge graph (KG). Many studies have been conducted with KG such as  \cite{ADMA1} and recently research related to KG \cite{ADMA2,nguyen2021judo,nguyen2020factcatch}.

However, understanding the performance implications of these techniques is a challenging problem. While each of them has distinct performance characteristics, their performance evaluations often lack diversity in domains, query languages, natural languages and datasets.
First, many developed datasets contain questions expressed in plain language but answers are specific to a KB format such as DBPedia.
Additionally, the number of questions in each of the  existing benchmarks is substantially different, rendering it difficult for users to select a practical guideline. 
Second, there is a lack diversity in terms of query languages. 
Another issue is the application domain. Most KBQA systems assume the generalization over different domains, but only a few domain-specific datasets are evaluated~\cite{costa2018leveraging,nguyen2019user,nguyen2020entity,nguyen2021structural,nguyen2019maximal}.

More precisely, the salient contributions of our benchmark are highlighted as follows:
\begin{compactitem}
    


    \item \textbf{Reproducible benchmarking:} 
    We present the first large-scale replicable benchmarking framework for comparing KBQA techniques. Additionally, it can be applied to novel methods proposed in future studies. 
    Our findings are reliable and reproducible, and the source code is publicly available at {\url{https://github.com/tamlhp/kbqa}}.
	
	\item \textbf{Surprising findings:} While some of our experimental findings confirm the state-of-the-art, we demonstrate the surprising superiority of methods using both SPARQL and SQL query languages. Additionally, we discovered that there are differences in their effectiveness when compared together.

	\item \textbf{New KGQA dataset:} We present COVID-KGQA, a new knowledge graph question answering benchmarking corpus that includes the most questions about COVID-19. COVID-KGQA includes more than 1000 questions with different types.

    \item \textbf{Performance guideline.} We provide valuable performance guidelines to academics working on KBQA in terms of benchmarks and QA systems selection. Alternatively, we suggest some future attempts to improve the performance.

\end{compactitem}

The remainder is organized as follows. 
We discuss in \autoref{sec:methodology} the problem setting, our benchmarking procedure, and the most representative approaches in KBQA. \autoref{sec:setup} introduces the setup used for our benchmark, including the analysis of KBs, datasets, metrics, and evaluation procedures. \autoref{sec:exp} reports the experimental results. A discussion of practical guidelines and conclusion are provided in \autoref{sec:con}. 

\section{Methodology}
\label{sec:methodology}

\subsection{Problem Setting}
\label{sec:ProbStat}

\begin{figure}[!h]
    \vspace{-1em}
    \centering
    \includegraphics[width=0.6\linewidth]{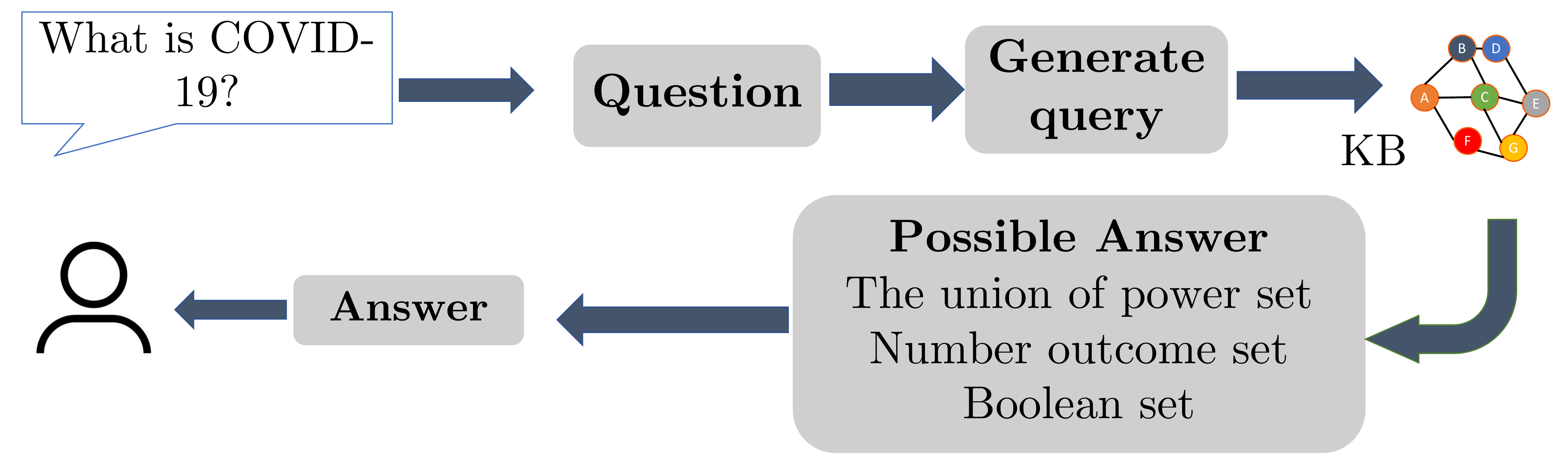}
    \caption{Question answering over knowledge bases.}
    \label{fig:problemStatement}
     \vspace{-2em}
\end{figure}

 Question Answering over Knowledge Bases (KBQA) is the term used for the task of retrieving the answer from executing query matching with natural language questions over a knowledge base as follow. Formally, we can define the task of KBQA as follow. Let $KB$ be a knowledge base, Q is a question, $q$ is the matching query and $A$ is an answer extracted by matching query $q$ executed from given question $Q$ over $KB$. According to \cite{chakraborty2021introduction}, the set of all possible answers can be in the form as follows. The first is the union of the power set $P$ of entities $E$ and literals $L$ in $KB$: $P (E \cup L)$. The second is the number of outcomes set for all potential functions of aggregation $F: P (E \cup L) \mapsto \mathbb{R} $. The third is a Boolean set of answers for yes/no questions. \autoref{fig:problemStatement} illustrates the general process of question answering over knowledge bases.

\subsection{KBQA Approaches}
\label{sec:techniques}

In recent years, the number of off-the-shelf methods for KBQA in a variety of applications has increased. It is fascinating to compare and evaluate them so that users can make informed choices.
KBQA can be divided into four primary strategies: (i) \emph{embedding-based} — converts a query into a logic form, which is then executed against KBs to discover the appropriate responses; (ii) \emph{subgraph matching} — constructs the query subgraph using a semantic tree; (iii) \emph{template-based} — converts user utterances into structured questions through the use of semantic parsing; and (iv) \emph{context-based}. This will be followed by a discussion of the concept underlying these approaches and some representative systems.

\subsubsection{Embedding methods}
\label{sec:embedding}
Embedding techniques take advantages of semantic and syntactic information included in a question as well as a database schema to produce a SQL logic form (parsing tree).

\sstitle{TREQS} 
The underlying concept of Translate-Edit Model for Question-to-SQL (TREQS) \cite{wang2020text} is to convert healthcare-related questions posed by physicians into database queries, which are then used to obtain the response from patient medical records. Given that questions may be linked to a single table or to many tables, and that keywords in the questions may not be correct owing to the fact that the questions are written in healthcare language, using the language generation method,
 this model is able to address the problems of general-purpose applications. Using a language generation model, this translate-edit model produces a query draft, which is subsequently edited in accordance with the table schema.

\sstitle{TREQS++}
TREQS++ or TREQS+Recover is the additional method with the added step as described below.
\emph{Recover Condition Values with Table Content:} This step is necessary because the use of the translate-edit model may not provide complete confidence that all of the queries will be executed since the condition values may not be precise in certain cases. With the goal of retrieving the precise condition values from the projected ones, this model employs a condition value recovery method in order to address this issue.

\subsubsection{Subgraph matching methods}

Subgraph matching constructs the query subgraph using a semantic tree, while others deviate significantly from this approach by building the subgraph from the entity.
A natural language phrase may have many interpretations, each of which corresponds to a different set of semantic elements in the knowledge graph. 
After the semantic tree is located, the semantic relationships must be extracted from it before the semantic query graph is constructed. 
\begin{figure}[!h]
\vspace{-2em}
    \centering
    \includegraphics[width=0.75\linewidth]{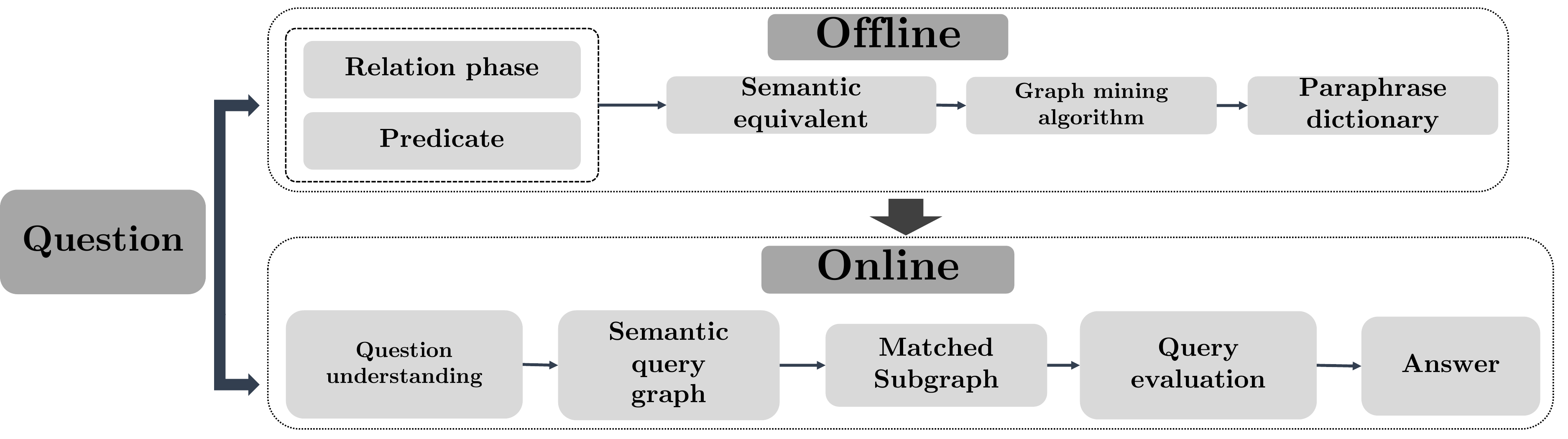}
    \caption{Overview of gAnswer model } 
    \label{fig:gAnswer}
\vspace{-2em}
\end{figure}

\sstitle{gAnswer}
gAnswer \cite{hu2017answering} is the most advanced subgraph matching method to convert natural language questions into query graphs that include semantic information. This model responds to natural language queries using a graph data-driven, offline and online solution. Using a graph mining technique, the semantic equivalence of relation terms and predicates is determined during the online phase. The discovered semantic equivalence is then included into a vocabulary of paraphrased phrases. The online part includes the question comprehension and assessment stages. During the question comprehension phase, a semantic query graph is constructed to capture the user's purpose by extracting semantic relations from the dependency tree of the natural language question using the previously constructed paraphrase dictionary. Then, a subgraph of the knowledge graph is chosen that fits the semantic query graph through subgraph isomorphism. The final result is determined by the chosen subgraph during the query assessment phase.


\subsubsection{Template-based methods}
The usage of templates is critical in question answering (QA) over knowledge graphs (KGs), where user utterances are converted into structured questions via the use of semantic parsing
. Using templates has the advantage of being traceable, and this may be used to create explanations for the users, so that they can understand why they get certain responses. 
\begin{figure}[!h]
\vspace{-1em}
    \centering
    \includegraphics[width=0.8\linewidth]{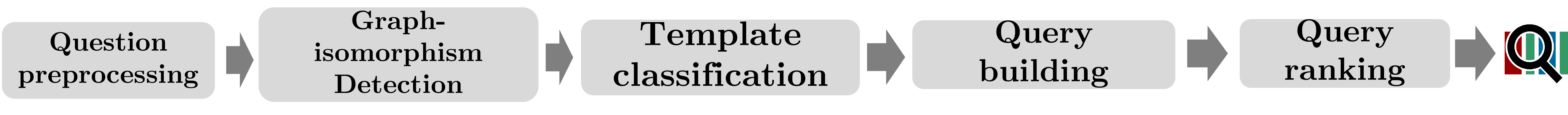}
    \caption{Overview of TeBaQA model } 
    \label{fig:tebaqa}
    \vspace{-2em}
\end{figure}

\sstitle{TeBaQA}
TeBaQA~\cite{vollmers2021knowledge} is the most advanced template-based technique.
First, all questions undergo (1) \emph{ Preprocessing} to eliminate semantically unnecessary terms and provide a meaningful collection of n-grams. By evaluating the underlying graph structure for graph isomorphisms, (2)\emph{the Graph-Isomorphism Detection and Template Classification} phase trains a classifier based on a natural language question and a SPARQL query using the training sets. The key assumption is that structurally comparable SPARQL searches correspond to questions with similar syntax. A query is categorised into a sorted list of SPARQL templates at runtime. During (3) \emph{ Information Extraction}, TeBaQA collects all relevant information from the question, such as entities, relations, and classes, and identifies the response type according on a set of KG-independent indexes. The retrieved information is entered into the top templates, the SPARQL query type is selected, and query modifiers are applied during the (4) \emph{ Query Building} step. The conducted SPARQL queries are compared to the predicted response type. The following (5) \emph{Ranking} is determined by a mix of all facts, the natural language inquiry, and the returning responses.

\subsubsection{Context-based methods}
\label{sec:IRTechnique}
These methods attempt to comprehend questions from various perspectives, such as question analysis, classification of questions, answer path, context of answers, and type of answers. 

\begin{figure}[!h]
\vspace{-1em}
    \centering
    \includegraphics[width=0.85\linewidth]{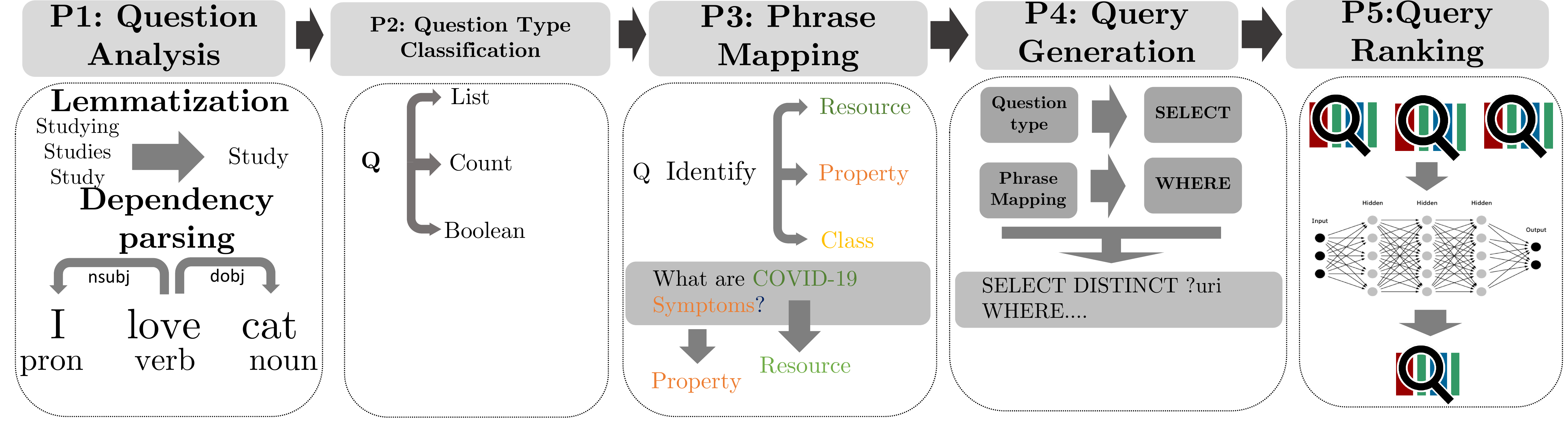}
    \caption{Overview of QAsparql model } 
    \label{fig:qasparql}
\vspace{-2em}
\end{figure}

\sstitle{QAsparql}
\label{sec:QAsparqlMethod}
QAsparql \cite{liang2021querying} is a model using five steps as listed below, to translate questions to SPARQL queries. \autoref{fig:qasparql} shows an overview of QAsparql model. 

At each stage, each individual software component separately solves a related job. First, the question analysis component processes the incoming question based only on syntactic characteristics. The question's phrases are then mapped to relevant resources and attributes in the underlying RDF knowledge network once the question's type has been determined. On the basis of the mapped resources and attributes, many SPARQL queries are built. A ranking model based on Tree-structured Long Short-Term Memory (Tree-LSTM) is used to order potential questions based on the closeness of their syntactic and semantic structure to the input query. Finally, results are given to the user by running the produced query against the knowledge network underpinning the system.

\sstitle{QAsparql* (Our proposal)}
While carrying out a resource and property mapping task (Phase 3 as described in \autoref{sec:QAsparqlMethod}) utilizing EARL~\cite{dubey2018earl}, we find that there is a significant difference between the two methods, as described in more detail below. In empirical evaluation, we have discovered that new algorithm produces superior results to previous one.

    
These methods are utilised to determine the number of hops and connections for each candidate node. This information is then sent to a classifier, which ranks and scores the features. After the creation of three distinct types of objects (Entity, Golden item, and the URI of each item), the next step is to convert them to a list. Consequently, we now have three separate listings. The List of Entities,  Golden Item, and URI are examples of lists. The primary purpose of this method is to identify which entity URI corresponds to the golden item URI.
The new algorithm performs the same function as the old algorithm, but with a different approach. During the first phase, it determines which items in the list correspond to the golden item and moves those that do not into a separate list known as the "not found list." It will then conduct a second search for entities, this time in surface form, to ensure that no entity is overlooked. After that, it will delete any non-included items and return a list of the items.

\subsection{Summary}

In summary, we use six representative systems — TREQS, TREQS++, gAnswer, TeBaQA, QAsparql and QAsparql* across domains, natural languages, and query languages.
 \autoref{tab:benchmark} compares the key characteristics of each technique studied in this benchmark.

\begin{table}[!h]
\centering
\vspace{-2em}
\caption{Characteristics comparison between KBQA techniques.}
\label{tab:benchmark}
\resizebox{0.85\textwidth}{!}{
\begin{tabular}{lcccccc} 
\toprule
\textbf{Technique} & \textbf{Features} & \textbf{Query language} & \textbf{Paradigm} & \textbf{Number of steps} & \textbf{Domain} & \textbf{Natural language}  \\ 
\midrule
TREQS              & Deep learning     & SQL, SPARQL             & Embedding         & 5                        & Single          & Single                     \\
TREQS++            & Deep learning     & SQL, SPARQL             & Embedding         & 6                        & Single          & Single                     \\
gAnswer            & Hand-crafted      & SPARQL                  & Subgraph matching & 4                        & Multiple        & Multiple                   \\
TeBaQA             & Hand-crafted      & SPARQL                  & Template-based    & 5                        & Multiple        & Multiple                   \\
QAsparql           & Hybrid            & SPARQL                  & Context-based     & 5                        & Multiple        & Multiple                   \\
QAsparql*          & Hybrid            & SPARQL                  & Context-based     & 5                        & Multiple        & Multiple                   \\
\bottomrule
\end{tabular}
}
\vspace{-2em}
\end{table}
\section{Experimental setup}
\label{sec:setup}

\subsubsection{Datasets}

We construct a sizable collection of benchmarking datasets.

\sstitle{Multi-domain datasets} Multi-domain attempts to improve performance by distributing it over several domains, and has been successfully used in a variety of areas.
\emph{Generic} are data gathered from a wide range of sources, including mathematics, physics, computer science, and a variety of other fields. 
	LC-QUAD \cite{trivedi2017lc} is Large-Scale Complex Question Answering Dataset. It consists of 5000 questions and answers pair with the intended SPARQL queries over knowledge base in DBPedia. 
\emph{Biomedical} is data about human health. 
	A large-scale healthcare Question-to-SQL dataset, MIMICSQL \cite{wang2020text}, was created by utilizing the publicly available real world Medical Information Mart for Intensive Care III (MIMIC III)
	dataset to generate 10,000 Question-to-SQL pairs. MIMICSQL* \cite{park2020knowledge} is the modified version of MIMICSQL which improves on the disadvantages of MIMICSQL as their tables are unnormalized and simpler than the tables used in actual hospitals. MIMICSPARQL is an SPARQL-based version of MIMICSQL*.

\sstitle{Multilingual datasets} Multilingual datasets are those that support multiple languages. With multilingual datasets, the performance of a model can be evaluated across a variety of languages, and the differences between each language can be determined. Beginning in 2011 and lasting until 2020, Question Answering over Linked Data (QALD) is a series of evaluation campaigns. The most recent version is QALD-9 \cite{ngomo20189th}, which contains 408 questions compiled and selected from previous tests and is available in eleven languages.

\sstitle{Our new benchmarking data (COVID-KGQA)}
In addition to the provided datasets, we create a bilingual dataset about COVID-19 to increase the diversity of domains and languages. Using the most recent version of DBPedia as a starting point, we compiled a corpus of more than one thousand question-answer pairs in two languages.




\subsubsection{Measurements}

We use the following evaluation metrics.

\sstitle{F1-score}
Let $Q$ denotes the number of questions in benchmark , $G$ is the number of answers processed by the system for given question $Q$, and $A$ is the number of corrected answers extracted by executing query $q$ from question $Q$ over knowledge graph $KG$. 
F1 is defined as 
$
F1 =\frac{1+\beta^2}{\beta^2*(1/P + 1/R)}
$
where $P=\frac{\mid A \mid }{\mid G \mid}$ and $R=\frac{\mid A \mid}{\mid Q \mid}$. $\beta$ weighs whether precision or recall is more important, and equal emphasis is given to both when $\beta$ = 1. 

\sstitle{Execution accuracy} 
Execution accuracy ($Acc_{EX}$) is computed as the accuracy of the response retrieved through SQL/SPARQL\cite{wang2020text}, 
$ Acc_{EX} = N_{EX}/N $, where N represents the total number of Question-SQL pairings in the MIMICSQL database and $N_{EX}$ represents the number of created SQL queries with the potential to provide accurate responses. Execution accuracy may also take into account questions formulated with invalid SQL queries but yielding valid query results. Therefore, we employ a different metric, Logical form accuracy, which eliminates the disadvantage of execution accuracy in the execution.

\sstitle{Logical form accuracy}
The Logical form accuracy ($Acc_{LF}$) is used for if a string match exists between a produced SQL query and a ground truth query~\cite{park2020knowledge}. In order to compute $Acc_{LF}$, we must compare the produced SQL/SPARQL with the true SQL/SPARQL, token by token. That is,
$
Acc_{LF} = N_{LF}/N 
$
where $N_{LF}$ counts how many requests are completely matched to the ground truth query. 

\sstitle{Computation time}
Another measure for KBQA methods is the amount of time required to complete a task.
\section{Results}
\label{sec:exp}




\subsection{End-to-end comparison}

\begin{figure}[!h]
\vspace{-1.0em}
  \centering
  \begin{subfigure}[b]{0.54\textwidth}
         \centering
         \includegraphics[width=1\linewidth]{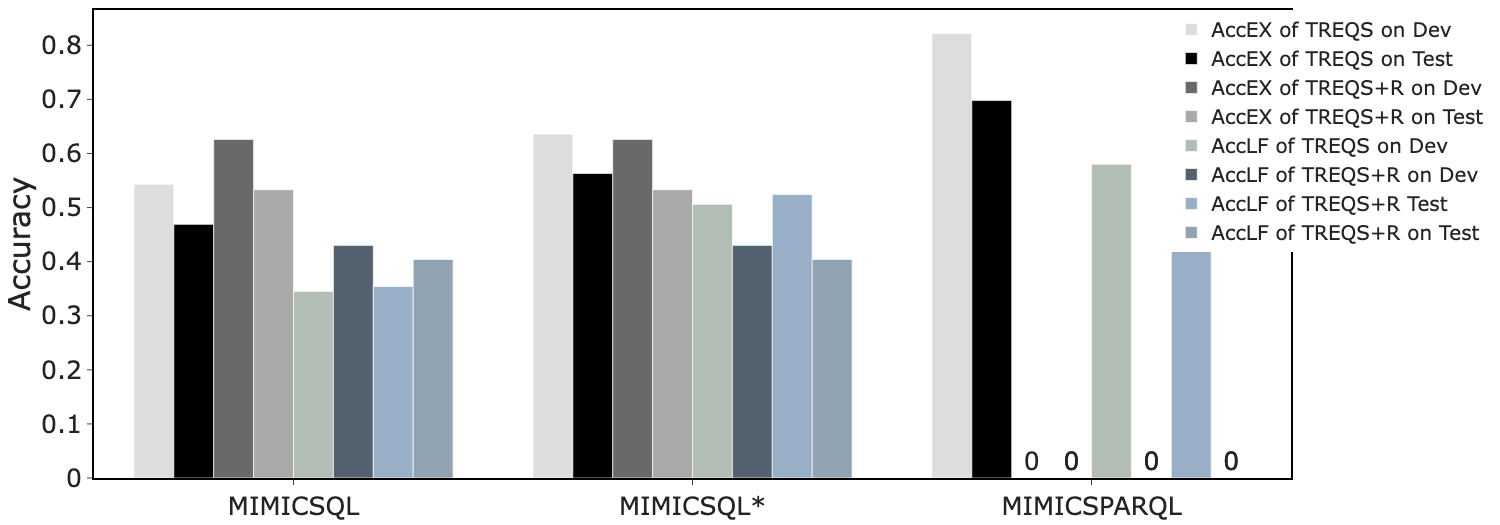}
  			\caption{On biomedical KBs}
  			\label{fig:mimicCompare}
     \end{subfigure}
  \begin{subfigure}[b]{0.44\textwidth}
         \centering
         \includegraphics[width=1\linewidth]{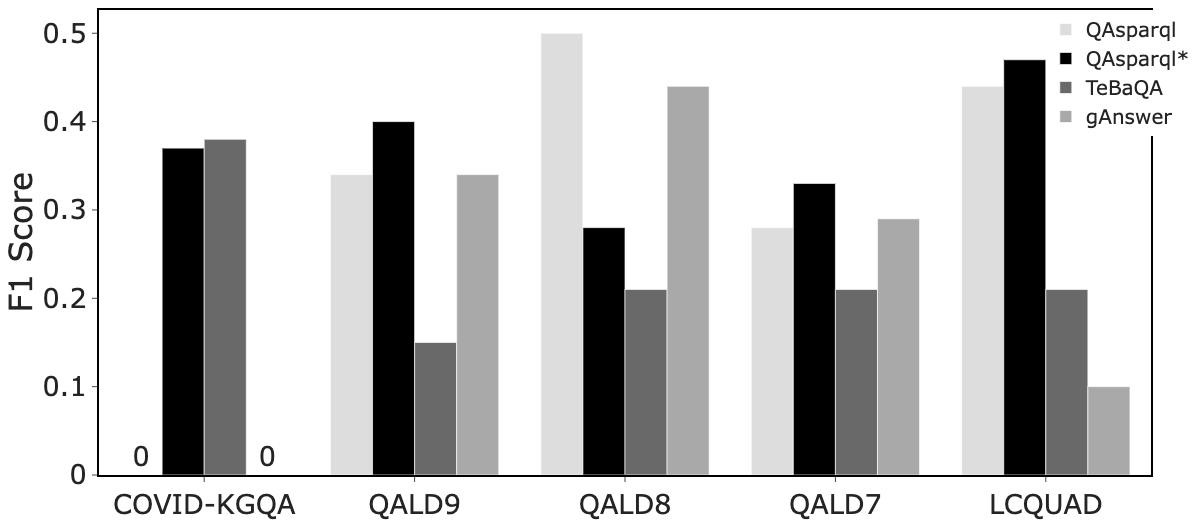}
    \caption{On generic KBs}
    \label{fig:GenericCompare}
     \end{subfigure}
     \vspace{-1em}
\caption{Accuracy comparison}
\vspace{-1em}
  \end{figure}

We provide empirical finding and examined the overall performance of knowledge bases in terms of answering questions on three biomedical datasets, four commonly used datasets, and one new dataset as part of this investigation. 
TREQS and TREQS++ are the most accurate among the top performers shown in \autoref{tab:Evaluation} on the first three biomedical datasets, with the highest accuracy on the fourth. TREQS++ achieves a better result on MIMICSQL, but its overall performance on the development and test datasets is inferior to that of TREQS on MIMICSQL*. Due to the design of TREQS++, which is to Recover Condition Values with Table Content, it is currently unable to perform with a knowledge graph and therefore cannot operate with MIMICSPARQL.
The figure \autoref{fig:mimicCompare} depicts the visualisation of our results on the first three biomedical datasets, while the figure \autoref{fig:mimicCompareTime} illustrates the average time required to answer 100 random questions. With the TREQS technique, $Acc_{EX}$ and $Acc_{LF}$ perform better on the development set than on the test set in the majority of instances. However, with TREQS, the outcomes are different; $Acc_{LF}$ performs better on the test set.

\begin{table}[!h]
\vspace{-2em}
\centering
\footnotesize
\caption{End-to-end comparison on biomedical datasets}
\label{tab:Evaluation}
\scalebox{0.7}{
\begin{tabular}{clccc} 
\toprule
\textbf{Dataset }                      & \multicolumn{1}{c}{\textbf{Method }} & \multicolumn{1}{l}{\textbf{$Acc_{EX}$}}      & \textbf{$Acc_{LF}$}                          & \multicolumn{1}{l}{\textbf{Time (s)}}                           \\ 
\toprule
\multirow{4}{*}{\textbf{MIMICSQL }}    & TREQS (Dev)                          & 0.543                                     & 0.345                                     & \multirow{4}{*}{\textcolor[rgb]{0.129,0.129,0.129}{0.161}}   \\
                                      & TREQS++ (Dev)                & {\textbf{0.626}}           & \textbf{{0.43}}            &                                                                 \\ 
\cline{2-4}
                                      & TREQS (Test)                         & 0.469                                     & 0.354                                     &                                                                 \\
                                      & TREQS++ (Test)               & {\textbf{0.533}}           & \textbf{{0.404}}           &                                                                 \\ 
\cmidrule[\heavyrulewidth]{1-5}
\multirow{4}{*}{\textbf{MIMICSQL* }}   & TREQS (Dev)                          & {\textbf{0.636}}           & {\textbf{0.506 }}          & \multirow{4}{*}{\textcolor[rgb]{0.129,0.129,0.129}{0.311}}   \\
                                      & TREQS++ (Dev)                & \textcolor[rgb]{0.129,0.129,0.129}{0.626} & \textcolor[rgb]{0.129,0.129,0.129}{0.43}  &                                                                 \\ 
\cline{2-4}
                                      & TREQS (Test)                         & {\textbf{0.563 }}          & {\textbf{0.524 }}          &                                                                 \\
                                      & TREQS++ (Test)               & \textcolor[rgb]{0.129,0.129,0.129}{0.533} & \textcolor[rgb]{0.129,0.129,0.129}{0.404} &                                                                 \\ 
\cmidrule[\heavyrulewidth]{1-5}
\multirow{2}{*}{\textbf{MIMICSPARQL }} & TREQS (Dev)                          & {\textbf{0.822}}           & 0.580                                     & \multirow{2}{*}{\textcolor[rgb]{0.129,0.129,0.129}{~0.25}}  \\
                                      & TREQS (Test)                         & 0.698                                     & \textbf{{0.641}}           &                                                                 \\
\bottomrule
\end{tabular}
}
\vspace{-2em}
\end{table}

%

\autoref{tab:Evaluation2} compares the four systems and their scores for the four metrics - Precision, Recall, F1 and Time used in our paper with five datasets (four generic dataset and one additional biomedical dataset) and \autoref{fig:GenericCompare} shows visualisation of them in terms of F1 . Our system has been chosen alongside two new systems that are up to date and scheduled to be launched in 2021. These systems performed outstandingly in the previous comparison \cite{liang2021querying}. 
The unique aspect of this discovery is that altering the entity and resource algorithm, as stated in \autoref{sec:IRTechnique}, improves performance over and beyond the default method used before. This is referred to as QAsparql* in order to differentiate it from the default. gAnswer is our next chosen system, which is state of the art in the QALD-9 challenge and is performed well on QALD-7, QALD-8, and LC-QUAD.

In terms of datasets, we select four existing datasets in addition to a large number of additional datasets that are split into the LC-QUAD and QALD series, which are the most frequently used datasets in the DBPedia. We choose QALD versions 7 to 9 from the QALD series since the target KG of these datasets are from version 2016, which is suited with a system that is nearly as current as the QALD series, and this will help these systems perform better. Our COVID-KGQA uses the latest version of DBPedia.

\begin{table}[!h]
\vspace{-2em}
\centering
\caption{Performance comparison on multilingual datasets (time in s)}
\label{tab:Evaluation2}
\resizebox{0.85\linewidth}{!}{
\begin{tabular}{lllllllllllllllllllll} 
\toprule
                  &  & \multicolumn{4}{c}{\textbf{\textbf{QAsparql}}}                                                                                        & \multicolumn{1}{c}{\textbf{}} & \multicolumn{4}{c}{\textbf{QAsparql*}}                                                                            &  & \multicolumn{4}{c}{\textbf{TeBaQA}}                                                                                                   & \multicolumn{1}{c}{\textbf{}} & \multicolumn{4}{c}{\textbf{gAnswer}}                                                                                                   \\
\textbf{Dataset}   &  & \multicolumn{1}{c}{\textbf{P}} & \multicolumn{1}{c}{\textbf{R}} & \multicolumn{1}{c}{\textbf{F1}} & \multicolumn{1}{c}{\textbf{Time}} & \multicolumn{1}{c}{\textbf{}} & \multicolumn{1}{c}{\textbf{P}} & \multicolumn{1}{c}{\textbf{R}} & \multicolumn{1}{c}{\textbf{F1}} & \textbf{Time} &  & \multicolumn{1}{c}{\textbf{P}} & \multicolumn{1}{c}{\textbf{R}} & \multicolumn{1}{c}{\textbf{F1}} & \multicolumn{1}{c}{\textbf{Time}} & \multicolumn{1}{c}{\textbf{}} & \multicolumn{1}{c}{\textbf{P}} & \multicolumn{1}{c}{\textbf{R}} & \multicolumn{1}{c}{\textbf{F1}} & \multicolumn{1}{c}{\textbf{Time}}  \\ 
\midrule
COVID-KGQA (Our) &  & 0.00                           & 0.00                           & 0.00                            & \textbf{0.26}                     &                               & 0.29                           & 0.52                           & 0.37                            & \textbf{2.69} &  & \textbf{0.38}                  & \textbf{0.38}                  & \textbf{0.38}                   & 13.2                              &                               & 0.00                           & 0.00                           & 0.00                            & 40.99                              \\
QALD-9 (Train)    &  & 0.4                            & 0.45                           & 0.42                            & 5.89                              &                               & 0.33                           & 0.71                           & 0.45                            & 32.37         &  & 0.17                           & 0.18                           & 0.16                            & 8.5                               &                               & 0.39                           & 0.43                           & 0.39                            & 13.03                              \\
QALD-9 (Test)     &  & 0.32                           & 0.36                           & 0.34                            & 9.56                              &                               & 0.3                            & 0.6                            & 0.4                             & 27.45         &  & 0.16                           & 0.15                           & 0.15                            & 8.15                              &                               & 0.33                           & 0.37                           & 0.34                            & 13.30                              \\
QALD-8 (Train)    &  & 0.35                           & 0.49                           & 0.41                            & 5.16                              &                               & 0.31                           & 0.67                           & 0.42                            & 7.9           &  & 0.21                           & 0.22                           & 0.21                            & 7.52                              &                               & \textbf{0.45}                  & 0.51                           & \textbf{0.46}                   & 12.93                              \\
QALD-8 (Test)     &  & \textbf{0.57}                  & 0.44                           & 0.5                             & 3.56                              &                               & 0.2                            & 0.44                           & 0.28                            & 7.35          &  & 0.21                           & 0.22                           & 0.21                            & 8                                 &                               & 0.42                           & \textbf{0.54}                  & 0.44                            & \textbf{12.07}                     \\
QALD-7 (Train)    &  & 0.33                           & 0.55                           & 0.42                            & 6.42                              &                               & \textbf{0.38}                  & 0.77                           & \textbf{0.51}                   & 5.69          &  & 0.2                            & 0.21                           & 0.2                             & 7.92                              &                               & 0.43                           & 0.5                            & 0.44                            & 13.09                              \\
QALD-7 (Test)     &  & 0.21                           & 0.44                           & 0.28                            & 1.74                              &                               & 0.24                           & 0.5                            & 0.33                            & 7.3           &  & 0.22                           & 0.23                           & 0.21                            & \textbf{7.72}                     &                               & \textbf{0.29}                  & 0.35                           & 0.29                            & 13.04                              \\
LC-QUAD (Test)    &  & 0.34                           & \textbf{0.62}                  & \textbf{0.44}                   & 5.41                              &                               & 0.31                           & \textbf{0.98}                  & 0.47                            & 13.75         &  & 0.21                           & 0.22                           & 0.21                            & 16.45                             &                               & 0.09                           & 0.11                           & 0.1                             & 60.45                              \\
\bottomrule
\end{tabular}}
\vspace{-1em}
\end{table}

For the COVID-KGQA dataset, we evaluated the effectiveness of the existing system with our new benchmark corpus in COVID-19 using four distinct systems. gAnswer is deployed via gStore with version 2016-10 of DBPedia as the guidance for gAnswer systems, whereas we conduct experiments with the most recent version of DBPedia for other systems. In this version, the knowledge graph contains no COVID-19-related information. Consequently, gAnswer achieves the worst performance in our corpus for both F1 and Time. On the other hand, TeBaQA and QAsparql have the best performance; while TeBaQA outperforms QAsparql 0.01 in F1, QAsparql outperforms TeBaQA in some fields, as discussed in \autoref{sec:QuestionQuantity}. TeBaQA has the best performance on our new benchmarking dataset, whereas this system performs poorly on other datasets with a performance range of only 0.15 to 0.21. Possible causes include the fact that QALD is a more difficult benchmark with a large number of questions requiring complex queries with more than two triples to answer.
A deeper examination revealed that QALD-9 questions frequently required sophisticated templates that were either absent from the training questions or provided very limited assistance. The template-based approach in TeBaQA is inferior to the Context-based and Subgraph matching techniques, with the exception of LC-QUAD. QAsparql*, our new finding approach, achieves the best performance in almost all corpora, with the exception of COVID-KGQA (with a length of less than approximately 0.01) and QALD-8. This demonstrates that enhancing the algorithm for entity matching can improve the system's performance.

\subsection{Running time}

\begin{figure}[!h]
    \vspace{-1em}
    \centering
    \includegraphics[width=0.45\linewidth]{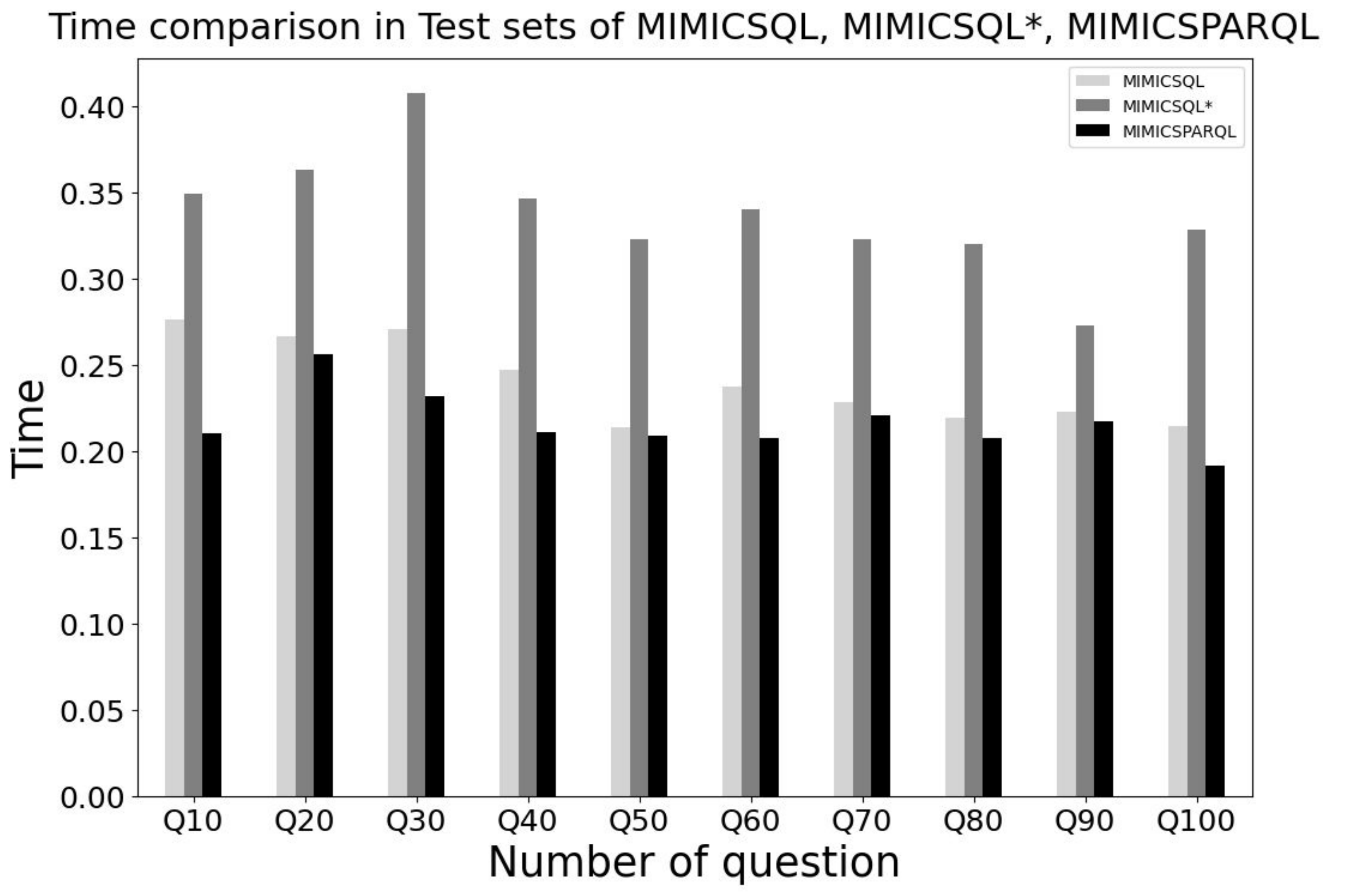}
    \includegraphics[width=0.45\linewidth]{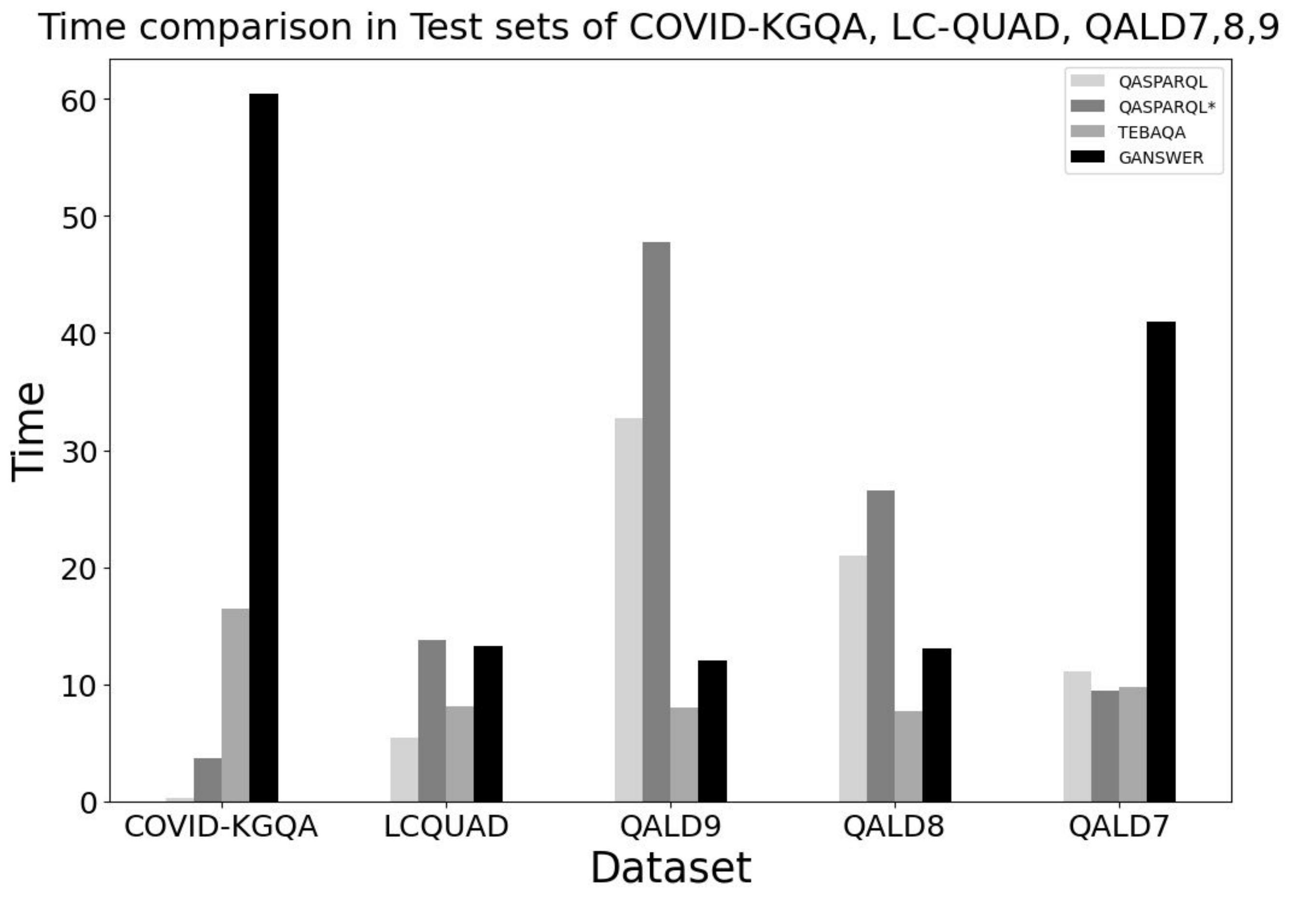}
    \caption{End to end time (s) comparison}
    \label{fig:mimicCompareTime}
    \vspace{-2em}
\end{figure}

\autoref{fig:mimicCompareTime} displays the experimental results. In the first three datasets, 100 questions are selected at random and their time performance is compared.
In comparison to two other corpora, MIMICSPARQL with a knowledge graph approach achieves the best results overall. Compared to MIMICSPARQL, MIMICSQL* queries are nearly twice as long. A single JOIN in SQL requires 11 tokens (including the operators "=" and "."), whereas a single hop in SPARQL requires only three tokens (i.e., subject, predicate, and object). An additional distinction between SQL and SPARQL is that the model must comprehend the hierarchy between a table and its columns and the relationships between tables. SQL and SPARQL differ syntactically due to this inherent distinction between relational tables and a knowledge graph in terms of how information is linked (joining multiple tables versus hopping across triples). Consequently, the graph-based method performs significantly better than the table-based method. \cite{park2020knowledge}.
We compare the execution times of four systems across five additional datasets. We evaluate every question in the test set of five corpora. gAnswer reaches its highest point in COVID-KGQA and becomes the fastest time overall. The explanation is summarised as follows: gStore is used to install gAnswer with the default knowledge graph version set to 2016-10. However, our dataset is about COVID-19, which was introduced in 2019, so gAnswer takes longer to query but returns no results.

\subsection{Influence of Question Taxonomy}
\label{sec:QuestionQuantity} 

\begin{figure*}[!h]
  \vspace{-1em}
\centering
     \begin{subfigure}[b]{0.49\textwidth}
         \centering
         \includegraphics[width=1\linewidth]{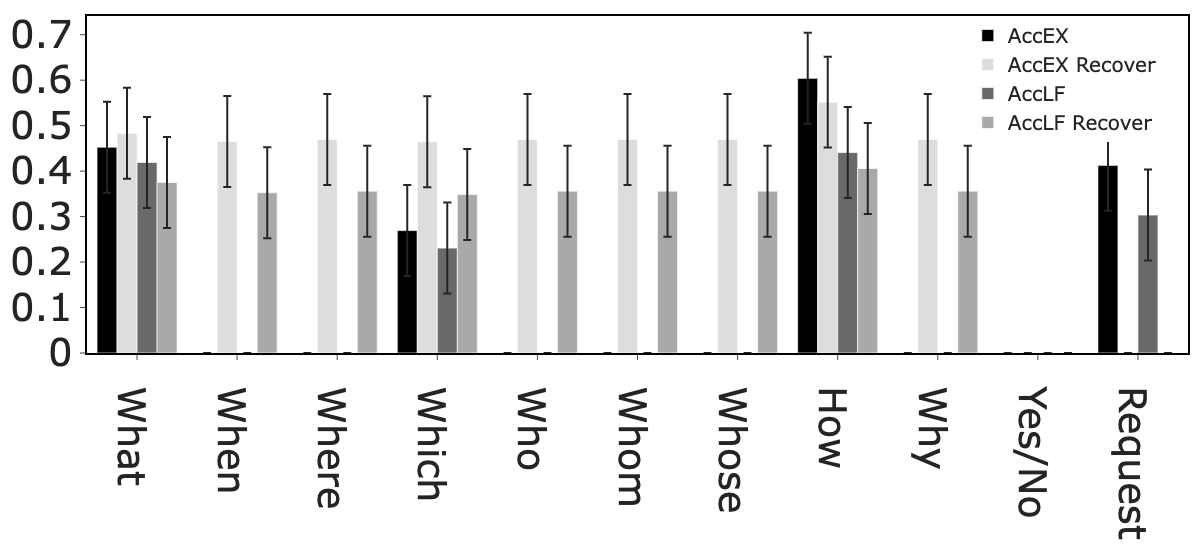} 
         \caption{MIMICSQL}
         \label{fig:QuestionTypeMIMICSQL}
     \end{subfigure}
     \begin{subfigure}[b]{0.49\textwidth}
         \centering
         \includegraphics[width=1\linewidth]{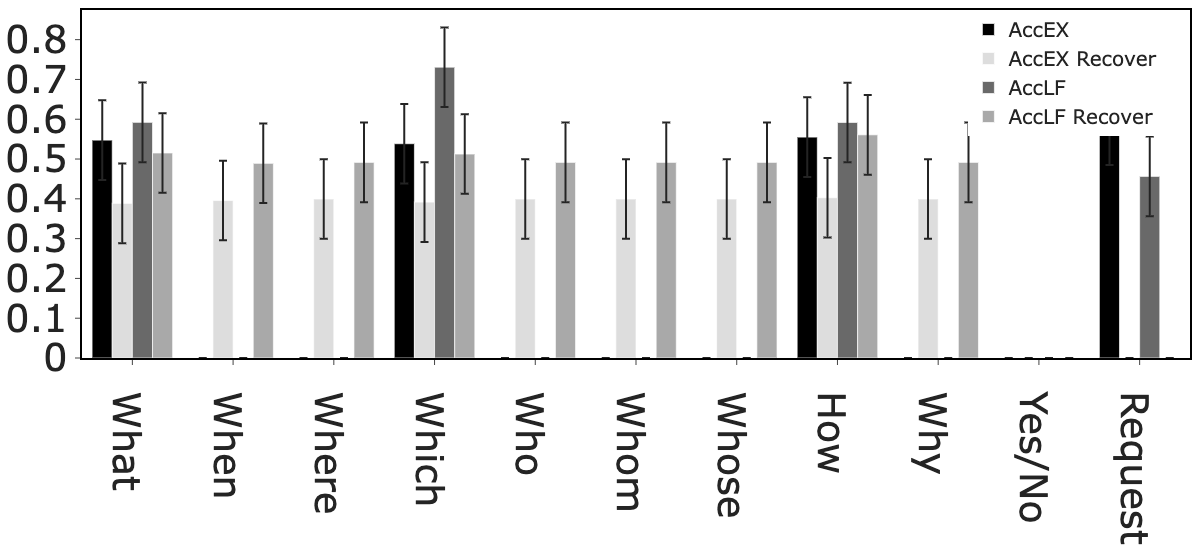} 
         \caption{MIMICSQL*}
         \label{fig:QuestionTypeMIMICSTAR}
     \end{subfigure}
     \begin{subfigure}[b]{0.49\textwidth}
         \centering
         \includegraphics[width=1\linewidth]{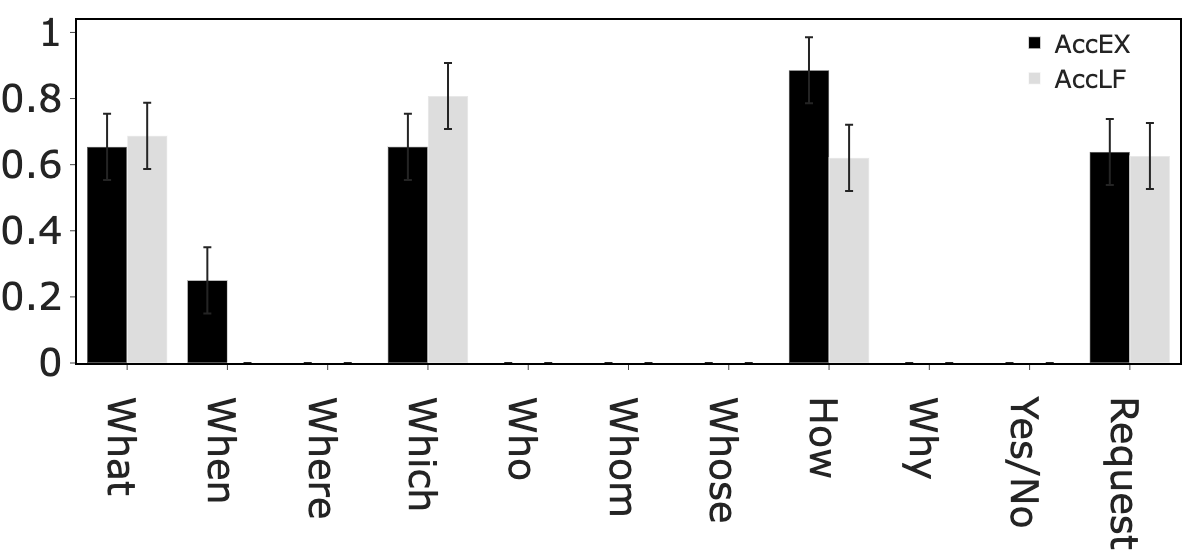} 
         \caption{MIMICSPARQL}
         \label{fig:QuestionTypeMIMICSPARQL}
     \end{subfigure}
     \begin{subfigure}[b]{0.49\textwidth}
         \centering
         \includegraphics[width=1\linewidth]{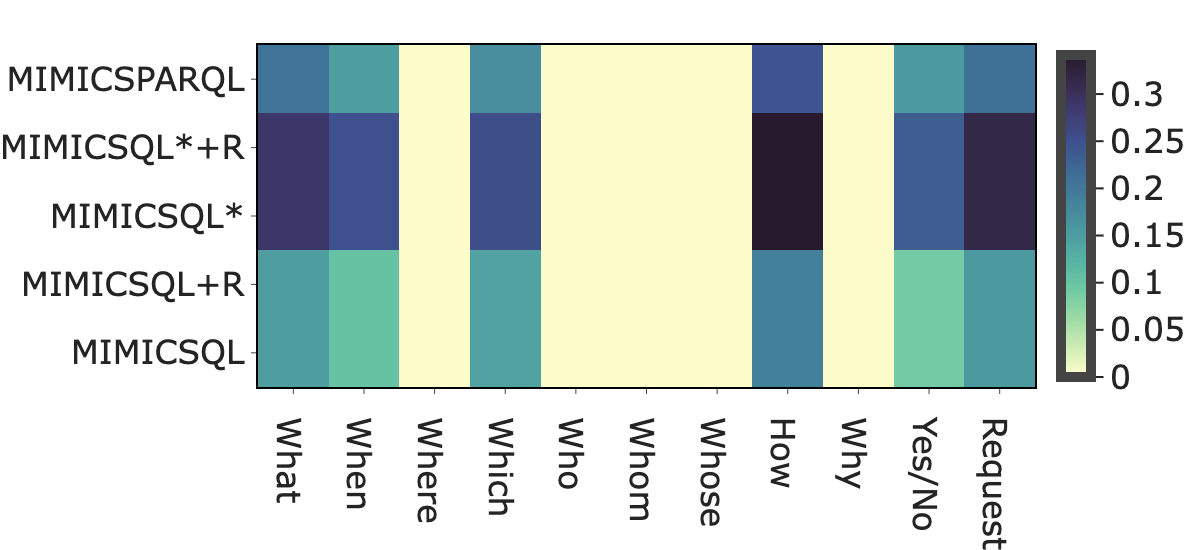} 
         \caption{Time}
         \label{fig:QuestionTypeMIMICTime}
     \end{subfigure}
        \caption{Comparison on MIMICSQL, MIMICSQL*, MIMICSPARQL with the influence of question type, the underscores represent zeroes}
        \label{fig:QuestionTypeMIMIC}
        \vspace{-1em}
 \end{figure*}

\begin{figure*}[!h]
\centering
     \begin{subfigure}[b]{0.5\textwidth}
         \centering
         \includegraphics[width=1\linewidth]{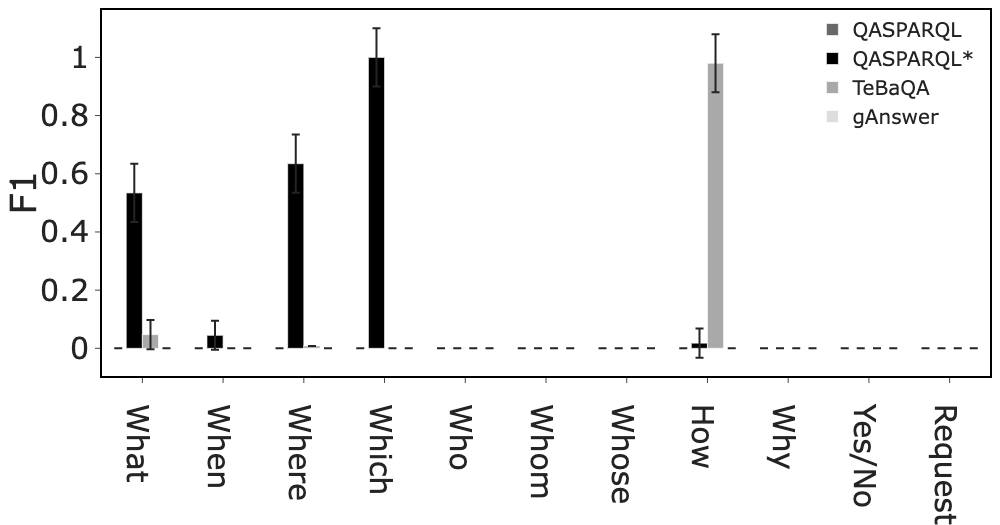}
         \caption{F1}
         \label{fig:QuestionTypeCOVIDF1}
     \end{subfigure}
     \begin{subfigure}[b]{0.4\textwidth}
         \centering
         \includegraphics[width=1\linewidth]{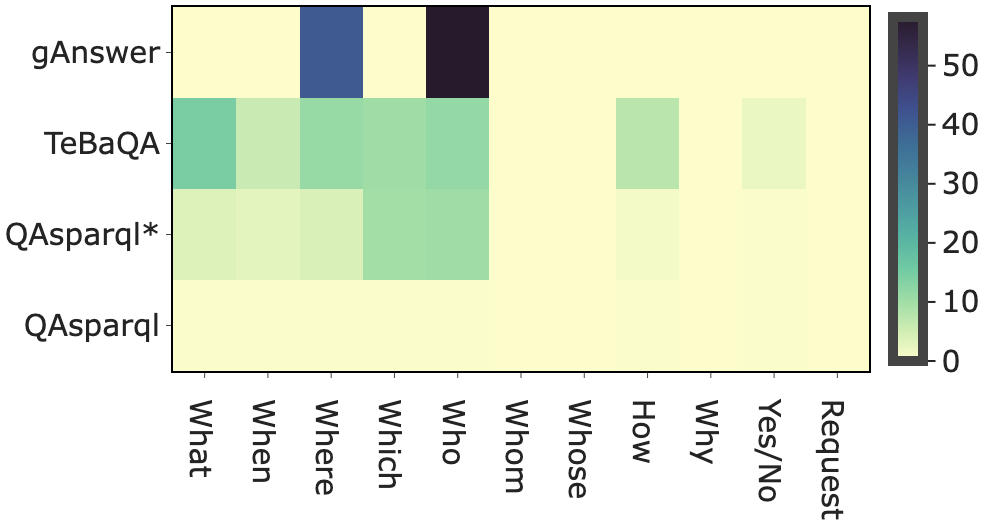}
         \caption{Time}
         \label{fig:QuestionTypeCOVIDTime}
     \end{subfigure}
        \caption{Comparison on COVID-KGQA with the influence of question type, the underscores represent zeroes}
        \label{fig:QuestionTypeCOVID}
 \end{figure*}

\begin{figure*}[!h]
\centering
     \begin{subfigure}[b]{0.54\textwidth}
         \centering
         \includegraphics[width=1\linewidth]{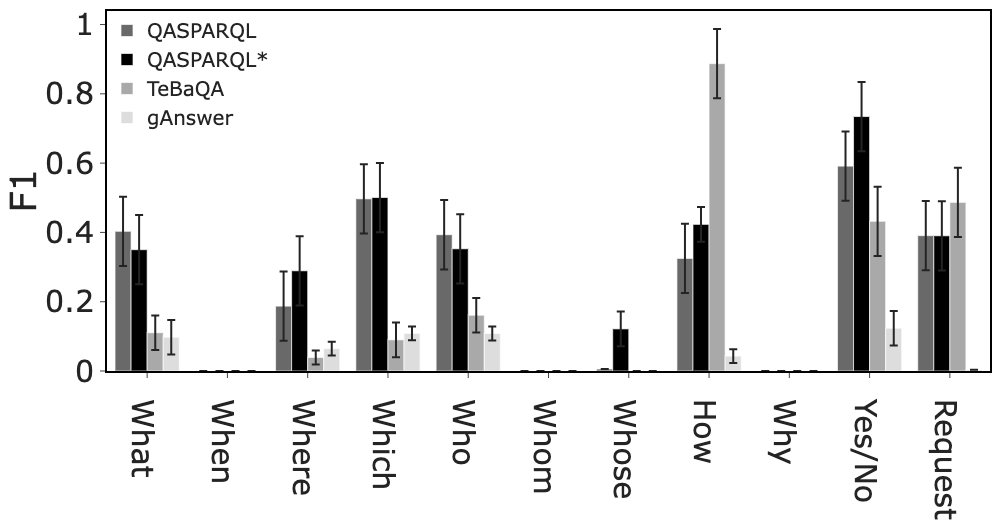} 
         \caption{F1}
         \label{fig:QuestionTypeLCQUADF1}
     \end{subfigure}
     \begin{subfigure}[b]{0.40\textwidth}
         \centering
         \includegraphics[width=1\linewidth]{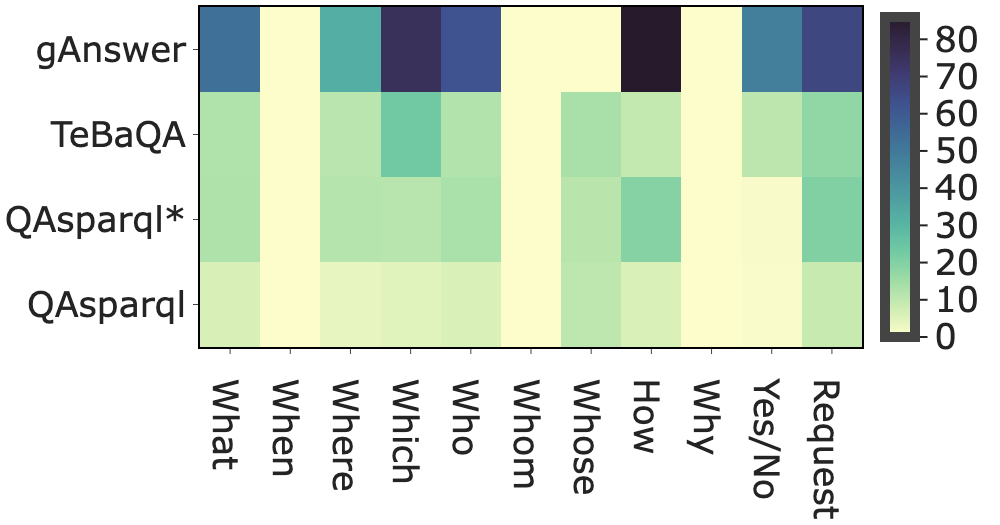}
         \caption{Time}
         \label{fig:QuestionTypeLCQUADTime}
     \end{subfigure}
       \caption{Comparison on LCQUAD with the influence of question type, the underscores represent zeroes}
    \label{fig:QuestionTypeLCQUAD}
 \end{figure*}

 Over the course of this investigation, the impacts of question type on question answering are investigated using knowledge graph methods. From these experiments, we conduct our studies on test set of LCQUAD and COVID-KGQA because the quantity of question of them is the highest and almost the same (1000 questions). \autoref{fig:QuestionTypeMIMIC}  depicts the results of MIMIC datasets, and \autoref{fig:QuestionTypeCOVID}, \autoref{fig:QuestionTypeLCQUAD}, which are conducted with four systems in COVID-KGQA and LCQUAD, as stated in \autoref{sec:QuestionQuantity}. According to \cite{pomerantz2005linguistic}, the Wh-question taxonomy comprises six types: 'Who', 'What', 'When', 'Where', 'Why', and 'How'. Additionally, throughout the data analysis process, we discover that the dataset contains other types such as 'Whose' and 'Whom'. On the other hand, certain inquiries are not Wh-questions, such as 'Yes/No' questions, which contain phrases starting with the words, i.e., 'Can', 'May', 'Am', 'Is', 'Are', 'Was', 'Were', 'Will', 'Do', and some request questions, such as 'Count', 'Provide', 'Give', 'Tell', 'Specify'. As a result, we undertake this experiment using eleven distinct sorts of questions.
 For the MIMICSQL dataset, \autoref{fig:QuestionTypeMIMICSQL} depicts the $Acc_{EX}$, $Acc_{LF}$, and Time of TREQS and TREQS++ for eleven question categories, with regard to each sort of query. The plot depicts the $Acc_{EX}$, $Acc_{EX}$ with TREQS++, and Time for each question type across the whole test set in this dataset. It is apparent that TREQS++ outperformed TREQS for all question categories on this dataset in $Acc_{EX}$ when using the Question-to-SQL techniques TREQS and TREQS++. $Acc_{LF}$ outperforms all other variables except What, How, and Request. Similarly, MIMICSQL* and MIMICSPARQL results are depicted in \autoref{fig:QuestionTypeMIMICSTAR} and \autoref{fig:QuestionTypeMIMICSPARQL}.
 Regarding the F1 in COVID-KGQA and LCQUAD results, these are illustrated in \autoref{fig:QuestionTypeCOVID} and \autoref{fig:QuestionTypeLCQUAD}. With respect to specific question categories, TeBaQA surpassed both of them. For COVID-KGQA, QAsparql* outperforms in the 4 categories of questions, including 'What', 'When', 'Where', and 'Which', whereas TeBaQA hits a high in the category of 'How'. 


%

 \begin{figure}[!h]

    \centering
    \includegraphics[width=0.8\linewidth]{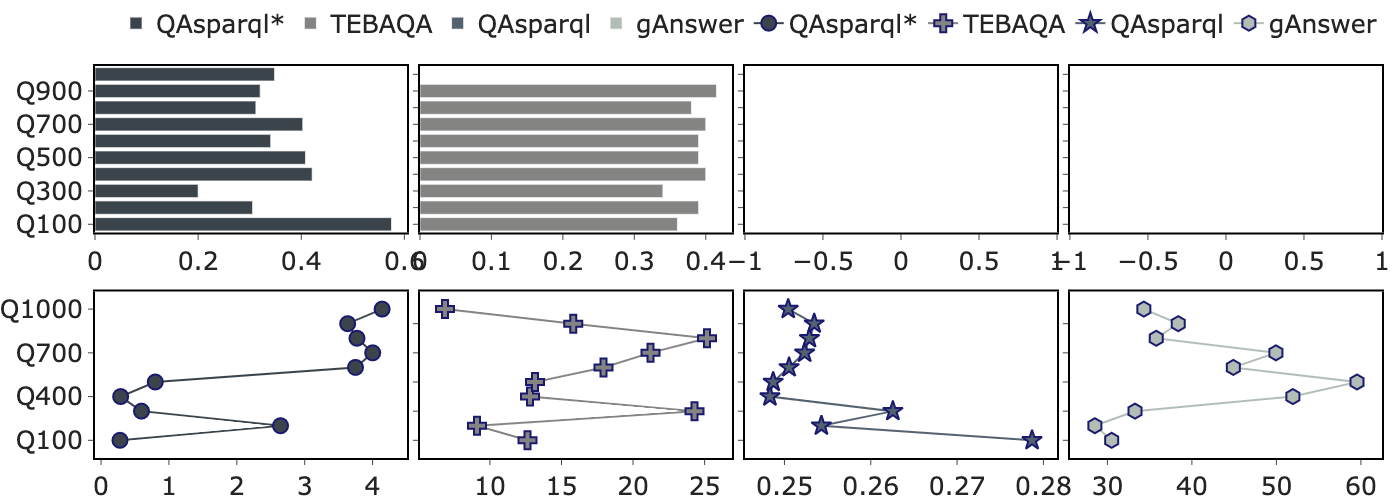} 
    \caption{Comparison on COVID-KGQA with the influence of question type from test set}
    \label{fig:QuestionNumberCOVID}

\end{figure}

\subsection{Effects of Quantity of Questions}

 \begin{figure}[!h]

    \centering
    \includegraphics[width=0.8\linewidth]{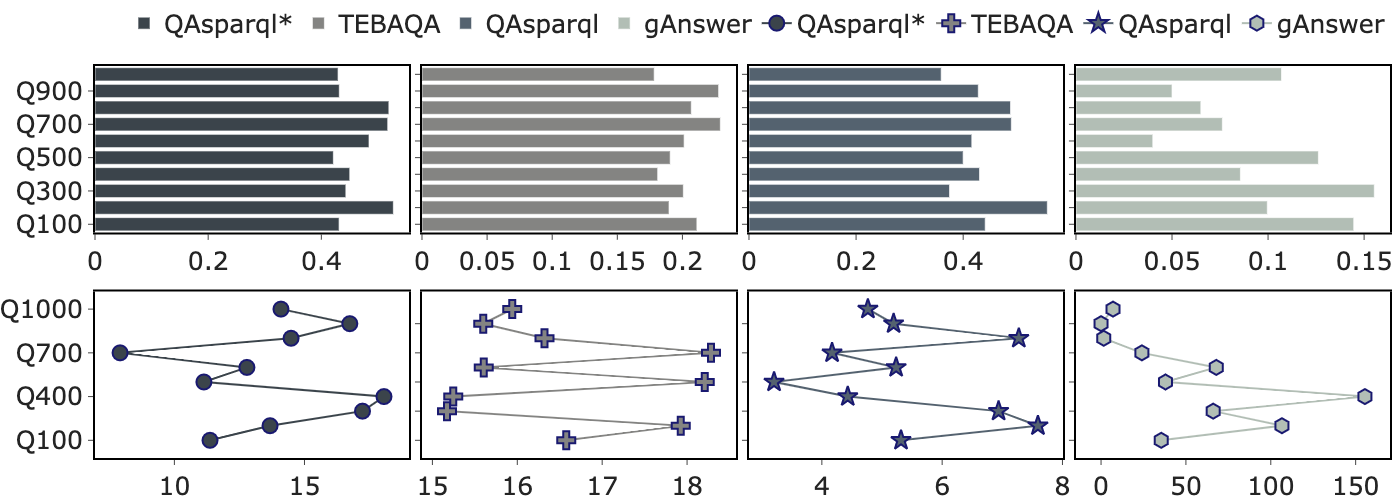} 
    \caption{F1 and Time comparison on LCQUAD with the influence of question quantity from test set }
    \label{fig:QuestionNumberLCQUAD}

\end{figure}

\begin{figure*}[!h]
\vspace{-1em}
\centering
     \begin{subfigure}[b]{1\textwidth}
         \centering
         \includegraphics[width=1\linewidth]{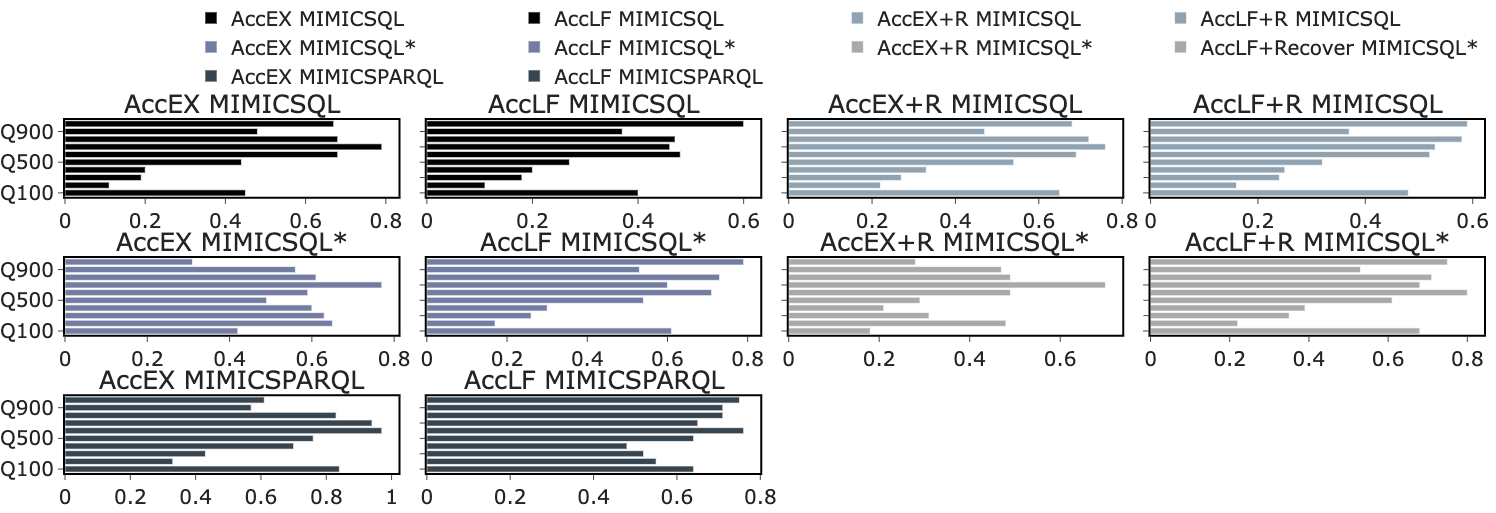}
         \caption{Accuracy}
         \label{fig:QuestionTypeCOVIDF1}
     \end{subfigure}
     \begin{subfigure}[b]{0.7\textwidth}
         \centering
         \includegraphics[width=1.0\linewidth]{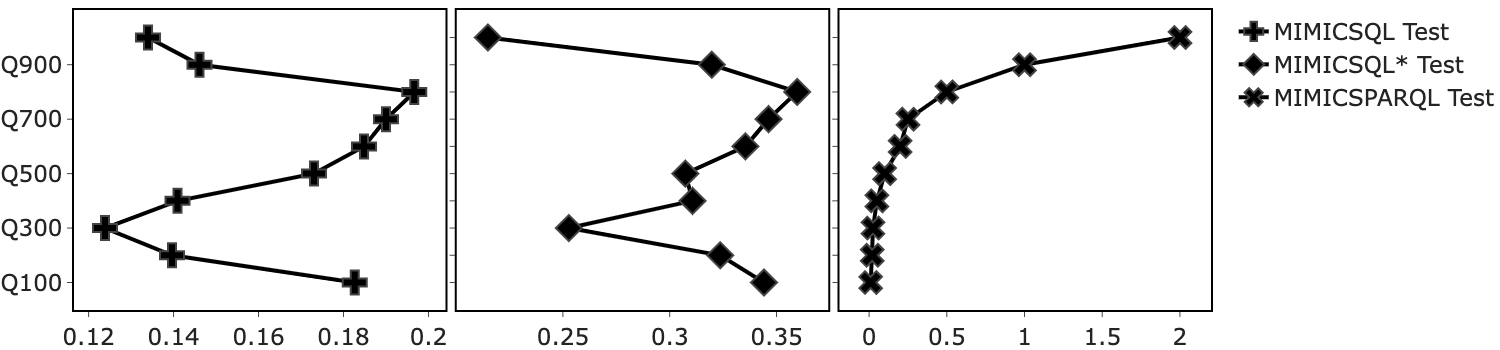}
         \caption{Time}
         \label{fig:QuestionTypeCOVIDTime}
     \end{subfigure}
        \caption{Effects of \#questions to Accuracy and Time on domain-specific datasets}
        \label{fig:mimicQuestionQuantity}
 \end{figure*}

The effects of next characteristic, the number of questions, are then examined. \autoref{fig:mimicQuestionQuantity} depicts the results of an experiment in which the quantity factor of the question is changed from each 100 question segments to 1000 questions of each dataset.
The accuracy of the test set improves significantly as the number of questions increases from 200 to 700, decreases from 800 to 900, and in the majority of instances $Acc_{EX}$ outperforms $Acc_{LF}$. The performance of both metrics typically peaks between the 500th and 700th question. Regarding Time in MIMICSQL, MIMICSQL*, and MIMICSPARQL, as depicted in \autoref{fig:mimicQuestionQuantity}, MIMICSQL and MIMICSQL* are quite comparable in terms of overall performance on the Test sets. 
The primary difference is that in MIMICSPARQL, the time for both questions is initially extremely short, but steadily increases until the 1000th question, where it reaches a maximum.

As shown in \autoref{fig:QuestionTypeCOVID} and \autoref{fig:QuestionTypeLCQUAD}, F1 in COVID-KGQA and LCQUAD. Regarding context-based techniques, it is evident that QAsparql* outperformed all other methods, with all questions scoring over 0.4 and 300 questions scoring over 0.5, and that it was surpassed by QAsparql* in terms of overall performance. In contrast, the gAnswer technique, which employs Subgraph Matching, had the lowest F1 scores of any technique. QAsparql* was superior to all other techniques and procedures. In terms of time, QAsparql achieved the highest performance, followed by QAsparql* and TeBaQA; however, gAnswer once again achieved the lowest performance in addition to F1 score.

\section{Conclusion}
\label{sec:con}

\sstitle{Performance guidelines}
To assist end users in selecting a suitable solution for a certain application need, we offer the following set of recommendations, which are based on our experimental findings.
\begin{itemize}

  \item In general, context-based methods, such as QAsparql and QAsparql*, are the most effective overall for both F1 and Time. 
The effectiveness of QAsparql decreased with question type. Despite its low computation time, TeBaQA struggles with question types and performs best with multiple criteria.

  \item In terms of $Acc_{EX}$ and $Acc_{LF}$, TREQS++ outperforms MIMICSQL in nearly all question types.
TREQS++ is ineffective in MIMICSPARQL, whereas TREQS is effective at $Acc_{EX}$ in How and Request questions. 
Regarding Time, QAsparql is the first runner up in nearly all kinds of questions, while QAsparql* is the second runner up in a negligible number of questions. 

  \item Regarding the effect of the number of questions on the performance of COVID-KGQA, we recommend avoiding the use of gAnswer and QAsparql, as these algorithms are inadequate in this regard. In terms of diversity, QAsparql* and TeBaQA are better. 

  \item Regarding multi-domain and multi-language, we advise avoiding the use of gAnswer and QAsparql, as these algorithms are ineffective in this regard. In terms of variety, QAsparql* and TeBaQA are both outstanding alternatives.

\end{itemize}

The findings are also summarized in \autoref{tbl:guideline}, where the best, the second-best and the worst techniques are shown for each percategory.

\begin{table}[!h]
\vspace{-1em}
\centering
\caption{Performance guideline} 
\label{tbl:guideline}
\scalebox{0.7}{
\begin{tabular}{lccc}
    \toprule 
     category  & winner & 1st runner-up & worst \\
    \midrule
    Overall &  QAsparql*  &  QAsparql*  & gAnswer \\
    Question type &  QAsparql*  & QAsparql  & gAnswer \\
    Question quantity &  QAsparql*  & QAsparql  & gAnswer \\
    $Acc_{EX}$ &  TREQS++  & TREQS  & TREQS \\
    $Acc_{LF}$ &  TREQS++  & TREQS  & TREQS \\
    Multi-domain & QAsparql*   & QAsparql*   & gAnswer \\
    Multi-language & QAsparql*   & QAsparql*   & gAnswer \\
    Cross-Query Language & TREQS  & TREQS  & TREQS++ \\
 \bottomrule
 \end{tabular}
}
\vspace{-1em}
\end{table}




\sstitle{Summary}
We compared KBQA systems from a variety of perspectives and discussed the ramifications of our findings.
Experiments were conducted on a wide range of topics, including simple subgraph matching, context-based, deep learning models, and others.
The results ultimately demonstrated the superiority of deep learning models when combined with a context-based approach, which produced the best results. In addition, as a result of this work, we create the COVID-19 large-scale KGQA dataset, which consisted of over a thousand questions.


\section*{Acknowledgement}
This research is funded by University of Information Technology, Vietnam National University HoChiMinh City under grant number D1-2022-25

%






\end{document}